\documentclass[conference]{IEEEtran}
\IEEEoverridecommandlockouts

\usepackage{mathtools}
\usepackage{lineno}
\modulolinenumbers[5]
\usepackage{comment}
\usepackage{graphicx}
\usepackage{amsfonts}
\usepackage{amsmath}
\usepackage{amssymb}
\usepackage{commath}
\usepackage{siunitx}

\newtheorem{rem}{Remark}

\newtheorem{assum}{Assumption}
\usepackage{placeins}
\usepackage{multirow}
\usepackage{array}
\usepackage[linesnumbered,lined,ruled]{algorithm2e}
\usepackage[misc,geometry]{ifsym}
\usepackage{mathtools,lipsum}
\usepackage{amsmath,amssymb,amsfonts}
\usepackage{algorithmic}
\usepackage{graphicx}
\usepackage{textcomp}
\usepackage{xcolor}
\def\BibTeX{{\rm B\kern-.05em{\sc i\kern-.025em b}\kern-.08em
    T\kern-.1667em\lower.7ex\hbox{E}\kern-.125emX}}

\DeclareMathOperator{\sign}{sign}

\usepackage[font=footnotesize]{subcaption}
\usepackage[font=footnotesize]{caption}
\usepackage{cuted}
\usepackage[capitalise]{cleveref}
\begin{document}

\title{A Feedback Motion Plan for Vehicles with Bounded Curvature Constraints\\
\thanks{This research was supported under NASA cooperative agreement number NND13AB04A.
}
}

\author{\IEEEauthorblockN{Giovanni Miraglia, Loyd R. Hook}
\IEEEauthorblockA{\textit{Electrical and Computer Engineering} \\
\textit{The University of Tulsa}\\
Tulsa, Oklahoma, USA \\
giovmi@utulsa.edu; loyd-hook@utulsa.edu}
}

\IEEEoverridecommandlockouts
\IEEEpubid{\makebox[\columnwidth]{978-1-7281-3885-5/19/\$31.00~\copyright2019 IEEE \hfill} \hspace{\columnsep}\makebox[\columnwidth]{ }}

\maketitle
\IEEEpubidadjcol

\begin{abstract}
The use of a feedback motion plan instead of the decoupled scheme consisting of separate plan and control phases can facilitate the task of proving the properties of an autonomous system. The advantage of using a feedback motion plan is the possibility to validate the whole plan offline before its execution, which means that trajectories having different initial states can be tested simultaneously. In this paper, we formulate a feedback motion plan based on the extension of the \emph{wavefront expansion} to the case of vehicles having bounded curvature. Additionally, the use of a transition function and a Gaussian filter limits undesired oscillations in the resultant trajectories. The method is suitable for both single goal missions and path following. The paper illustrates the algorithm for the generation of the plan and presents simulation data containing example trajectories and analysis of tuning parameters. Finally, future developments are discussed.
\end{abstract}


\section{Introduction}
Both industry and academia are working to define proper standards and certification strategies for autonomous transportation systems \cite{Hook2016}. The common requirement in these strategies is the ability to prove the robustness and safety of the autonomous system in different scenarios. To satisfy this general requirement for the navigation task specifically, it is crucial to prove that the autonomous system can recover safely after the occurrence of unpredicted events. This task is not trivial, especially if planning and control phases are decoupled. In this case, if an excessive deviation occurs it may be necessary to perform an on-line replan. Although in many instances this is acceptable, for applications such as those involving unmanned or autonomous air vehicle systems, computed paths are required to be pre-approved in order to assure compliance with regulations and safety constraints.
An alternative approach that allows for the simultaneous verification of a large number of paths \cite{Mira1910:Feedback}, is the use of a feedback motion plan \cite{LaValle2006}, where for each vehicle position and heading there is a specific pre-computed action that will be executed. This approach provides the ability to reject disturbances and recover from deviations thanks to the integration of planning and control. 

In our method, we generate a feedback motion plan for vehicles with bounded curvatures moving in the 2-D space. The method extends the \emph{wavefront expansion} algorithm \cite{Choset2005} to the case of vehicles having minimum turning radius constraints. The use of a transition function and a Gaussian filter limits the number of oscillations in the resulting path. Finally, since the formulation is based on the wavefront expansion, it allows for the presence of obstacles having non-convex shape.

The effectiveness of the method is demonstrated via simulations based on a 2-D kinematic model.

The paper is organized as follows. Section II provides a review of relevant related work. Section III provides the theoretical formulation of the problem, while Section IV illustrates the vector field generation process. In Section V,  simulations results are shown and discussed. Finally, we offer conclusions and discuss future work.

\section{Related Work}
The feedback motion plan presented in this paper consists of the generation of a pre-calculated vector field. In a vector field, a vector is assigned to each point on a manifold \cite{Triggs1993} to control a robot indicating the direction of state evolution. Vector fields have been exploited in both classical path planning and path following problems. The strength of methods based on vector fields is that it is possible to integrate path planning and robot control in the same approach. Furthermore, vector field guidance can lead the robot in the right direction even in the presence of small localization errors \cite{Goncalves2010}. A vector field can be generated employing the gradient $ \nabla f(x)$ of a potential function $ f:\mathbb{R}^{n}\rightarrow\mathbb{R} $. Given a potential function, the robot is guided using \textit{Gradient Descent} \cite{Choset2005}, where the robot takes a ``downhill'' path to reach the goal configuration. The main challenge is to guarantee the existence of a minimum only at the goal configuration. Many potential functions do not lead to complete path planners because of the existence of multiple local minima  \cite{Choset2005,Koren1991,LaValle2006}, which means that the robot can be ``trapped'' in wrong configurations. A solution to this problem is the use of harmonic functions \cite{Kim1992}. However, in this case, often it is difficult to find the appropriate harmonic function.

In addition to the local minima problem, another problem that affects methods based on vector fields is the presence of oscillation in the resulting trajectory. This phenomenon is often referred to as \textit{chattering} \cite{Choset2005}. To alleviate this problem Ren \textit{et al.} in \cite{Ren2006} propose the use of the modified \textit{Newton's method} (MNM), which is a technique used in optimization theory. They show that the use of MNM reduces the presence of oscillations regardless of the type of potential function. Nevertheless, the study presented in \cite{Ren2006} is restricted to the case of convex obstacles; therefore, it neglects the local minima problem that can arise with non-convex obstacles. 

Vector fields have also been employed for the guidance of nonholonomic robots \cite{Ren2008,Pathak2005,Conner2003}. In \cite{Lindemann2005} Lindemann and LaValle illustrate a method in which cell decomposition and local vector fields are employed. However in this case, instead of using the gradient of a potential function they directly build local vector fields and combine them smoothly using bump functions and \textit{generalized Voronoi diagram} (GVD).
  
The objective of the abovementioned guidance methods is to reach a single goal location; however, vector fields have also been employed in path following problems \cite{Goncalves2010}\cite{Miraglia2017} . In \cite{Nelson2007}, Nelson \textit{et al.} formulate a vector field for the guidance of a Small Unmanned Aerial Vehicle (UAV). In this case, the vectors represent the desired course inputs that must be fed to the attitude control loop. More specifically, the authors define continuous vector fields that can be used to follow both straight-line paths and circular arcs or orbits. In \cite{Jung2016} Jung \textit{et al.} extend the path following for straight-line paths and orbits to account for both angle and time arrival.

In all the above-mentioned methods, single goal mission and path following are addressed singularly. Furthermore, many of the methods consider only convex obstacles for single missions or neglect the presence of obstacles for path following. Instead, the method presented in this paper is versatile, because it can be used for both single goal missions and path following. Finally, the proposed method is general because it does not require a specific control law.

\section{Problem Formulation}
The feedback motion plan proposed in this paper is formulated for a specific class of constrained systems, which travel with a constant speed $v$ and have a limited turning rate $\omega$. The constraints to the vehicle's motion can be formulated in terms of minimum turning radius:
\begin{equation}\label{min_radius}
r_{\text{min}}=\frac{v}{\omega}
\end{equation}
A vehicle with this type of motion is usually referred to as \textit{Dubin's vehicle} \cite{Dubins1957}. In this case, we can use the following kinematic model:
\begin{equation}\label{state_transition}
\begin{split}
\dot{x}&=v\cos\theta\\
\dot{y}&=v\sin\theta\\
\dot{\theta}&= \omega
\end{split}
\end{equation}
where $(x,y)$ are the vehicle's coordinates in the workspace $W$, which is a bounded subset of $\mathbb{R}^{2}$.
In \cref{state_transition} the only control variable is the turning rate $ \omega$; therefore, it is the model of a first-order underactuated system. Path planning for this type of vehicle is often referred to as the \textit{Steering Problem} \cite{LaValle2001}.
In our formulation, the vehicle's configuration is $q=\left(x,y,\theta\right)$, with $\theta\in\left[0,2\pi\right)$, and the action space is $U=\{-\omega, 0, \omega\}$. With the identification $0\sim2\pi$, the configuration space is $C=\mathbb{R}^{2}\times\mathbb{S}^{1}$. The set $O\subset W$ is the region occupied by obstacles, therefore the free work space is $W_{\text{free}}=W/O$. The free configuration space is defined as $C_{\text{free}}=C/ C_{\text{obs}}$. Without loss of generality, we assume that $O$ is defined accounting for the actual vehicle's dimensions. In our discussion, the state space is $X=C$, thus the state transition equation $ \dot{x}=f\left(x,u\right)$ \footnote{With a little abuse of notation we use $x$ to indicate both the state and the state variable that corresponds to the coordinate in the Euclidean space} corresponds to \cref{state_transition}.
To account for minimum turning radius constraints, we define the \textit{buffer region} as follows:
\begin{equation}
W_{\text{buffer}}=\left\lbrace w \in W_{\text{free}} | \min\limits_{w_{o}\in O}\left\lbrace\norm{w-w_{o}}\right\rbrace\leq \alpha r_{\text{min}},\alpha\geq 2 \right\rbrace
\label{buff_region_def}
\end{equation}
where the scalar $\alpha$ defines the width of the buffer region. Finally, we define the region of \textit{safe start} (SS) as $W_{\text{SS}}=W_{\text{free}}/W_{\text{buffer}}$. The regions in the configuration space $C$ (and thus in the states space $X$) are obtained by extending the corresponding regions of the workspace along the $\theta$-axis.
\begin{figure}[b]
\centering
\begin{subfigure}[b]{0.20\textwidth}
\centering
\includegraphics[width=1\textwidth]{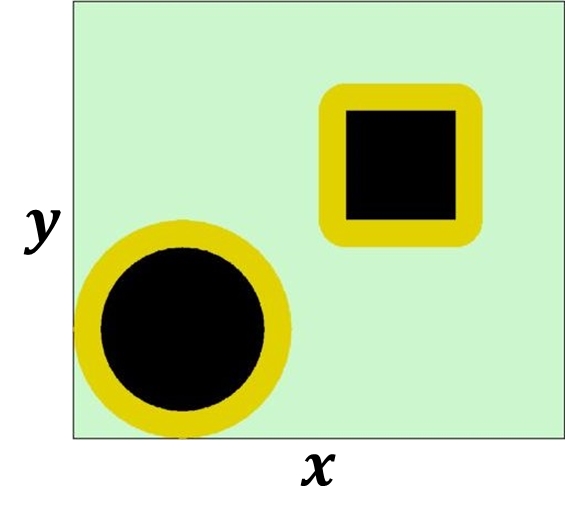}
\caption{ }
\label{workspace}
\end{subfigure} 
\begin{subfigure}[b]{0.28\textwidth}
\centering
\includegraphics[width=1\textwidth]{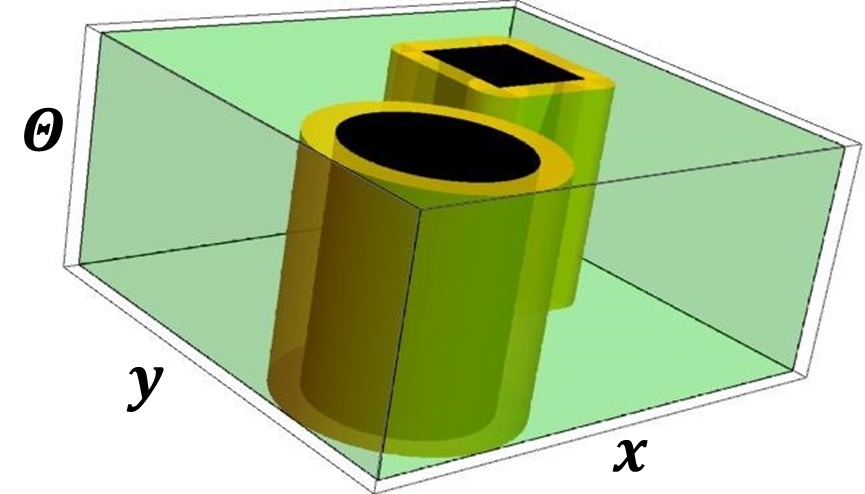}
\caption{ }\label{conf_space}
\end{subfigure}
\caption{Example of workspace (a) and  configuration space (b). Green indicates \textit{free} region, yellow indicates \textit{buffer} region and black indicates \textit{obstructed} region.}
\label{example_work_conf}
\end{figure}
Our objective is to find a feedback plan $\pi:X\rightarrow U$, which indicates the action $u\in U$ that must be taken to reach the goal set $X_{\text{G}}$ for every state $x\in X_{\text{free}}$. In our formulation we consider two scenarios:
\begin{enumerate}
\item single goal destination;
\item predefined path.
\end{enumerate}
In the first case, given the goal position $p_{\text{g}}=\left( x_{\text{g}},\small y_{\text{g}}\right)\in W_{\text{SS}}$ in the workspace $W$, the goal set is defined as follows:
\begin{equation}
X_{\text{G}}=\left\lbrace x \in X_{\text{free}}|\left( x-x_{g}\right)^{2}+\left( y-y_{g}\right)^{2}\leq \beta r_{\text{min}}, \beta\geq 2\right\rbrace
\label{single_goal_def}
\end{equation}
Therefore, the goal set is a cylinder having radius $\beta r_{\text{min}}$. This definition of the goal set is due to the vehicle's kinematics, which has a nonzero constant speed. 
In the second case, the goal set is defined as follows:
\begin{equation}
\begin{split}
X_{\text{G}}=\lbrace x \in X_{\text{free}}\vert( \exists x^{*} \in X_{\text{path}})[\vert x-x^{*}\vert < \delta_{x_{\text{max}}}\wedge\\\vert y-y^{*}\vert < \delta_{y_{\text{max}}}\wedge \vert \theta-\theta^{*}\vert < \delta_{\theta_{\text{max}}}]\rbrace
\end{split}
\label{goal_region_path_def}
\end{equation}
where $\delta_{*}$ is the maximum acceptable error for each state variable.
Henceforth, we will rely on the following assumption.
\begin{assum}
the goal region $X_{\text{G}}$ is entirely contained in the region of safe start $X_{\text{SS}}$, therefore   $X_{\text{G}}\cap X_{\text{buffer}}=\varnothing$ and $X_{\text{G}}\cap X_{\text{obs}}=\varnothing$.
\label{goal_in_ss}
\end{assum}

The feedback plan $\pi(x)$ is based on a discrete 2-D vector field, which can be defined as follows:
\begin{equation}
\vec{F}\left(x,y\right)=P\left(x,y\right)\hat{i}+Q\left(x,y\right)\hat{j}=\left[ P\left(x,y\right),Q\left(x,y\right)\right]^{T}
\label{vec_definition}
\end{equation}
Since the vehicle's velocity vector $\vec{v}$ has a constant magnitude, we can use a normalized vector field, in which the vector magnitudes are $1$. In this case, $\vec{F}\left(x,y\right)$ can be characterized using a single grid-map, in which the vector orientations $\angle\theta_{F}\left(x,y\right)$ is stored. Therefore, \cref{vec_definition} becomes:
\begin{equation}
\vec{F}\left(x,y\right)=\left[ \cos\angle\theta_{F}\left(x,y\right),\sin\angle\theta_{F}\left(x,y\right)\right]^{T}
\label{vec_normalized}
\end{equation}
We can now define the feedback plan as follows:
\begin{equation}
\pi\left(x\right)= \begin{dcases*}
\omega\sign \left(\Delta\theta_{\text{RF}}\left(x\right)\right), & $x\in X_{\text{buffer}}$\\
z\left(\Delta\theta_{RF}\left(x\right)\right), & $x\in X_{\text{SS}}$
\end{dcases*}
\label{plan_definition}
\end{equation}
where $x$ is the state, $\Delta\theta_{RF}\left(x\right)$ is the orientation error and $z(x)$ is the control law applied to correct the error. The formulation in  \cref{plan_definition} means that if the vehicle is inside the \emph{buffer region} it must turn with its minimum turning radius, while outside the \emph{buffer region} it can use any control law to correct the heading angle.

The heading error can be calculated as follows:
\begin{equation}
\Delta\theta_{RF}\left(x\right)= \begin{dcases*}
\frac{\cos^{-1}\left(\vec{v}\cdot\vec{F}\left(x,y\right)\right)}{\Vert\vec{v}\Vert}, & $\angle\theta_{v}\leq\angle\theta_{F}\left(x,y\right)$\\
-\frac{\cos^{-1}\left(\vec{v}\cdot\vec{F}\left(x,y\right)\right)}{\Vert\vec{v}\Vert}, &$\angle\theta_{v}>\angle\theta_{F}\left(x,y\right)$\\
\end{dcases*}
\label{orientation_error}
\end{equation}
Finally, we assume that the work space $W$ is represented with a 2-D bitmap array as follows:
\begin{equation}
\begin{split}
BM: W &\rightarrow \left\lbrace 1,0 \right\rbrace\\
x &\mapsto BM\left(x\right)
\end{split}
\end{equation}
where a $0$ indicates a cell of $W_{\text{free}}$, while a $1$ indicates a cell of $W_{\text{obs}}$. In our formulation, we will use the \textit{2-neighborhood} \cite{Barraquand1992}, often referred to as \textit{8-point connectivity} \cite{Choset2005,Petres2007}, which means that in the wavefront expansion, in addition to horizontal and vertical directions, the expansion is performed also in the diagonal directions.
\begin{rem}
In our formulation we discretize the $x$ and $y$ axes of the $C$-space, while the $\theta$-axis is continuous. Therefore, the  $C$-space is decomposed in rectangular boxes $\text{rec}_{i}=\left[ i\cdot \text{res}, \left(i+1\right)\cdot \text{res} \right)\times \left[ j\cdot \text{res}, \left(j+1\right)\cdot \text{res} \right)\times \left[0,2\pi \right)$, where $i\in\left\lbrace n\in \mathbb{N}\vert n<n_{x_{\text{max}}} \right\rbrace $ and $j\in\left\lbrace n\in \mathbb{N}\vert n<n_{y_{\text{max}}} \right\rbrace $ are the number of cells for respectively $x$-axis and $y$-axis, and \textit{res} is the resolution of the grid-map used for the workspace $W$.
\end{rem}

\section{Vector Field Generation}
\begin{figure*}[t]
\centering
\includegraphics[width=5in]{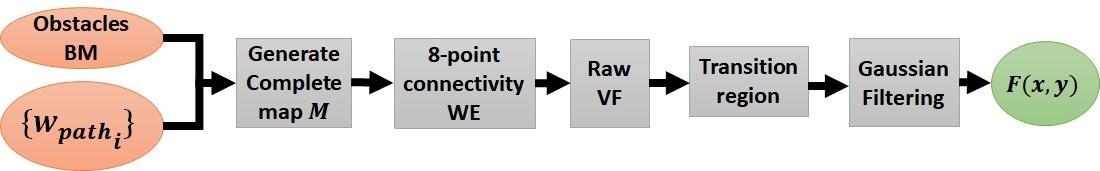}
\caption{Map generation process (WE: wavefront expansion, VF: vector field).}
\label{map_chain}
\end{figure*}
The process followed to generate the vector field is illustrated in \cref{map_chain}. The inputs are the bitmap $BM$ and the ordered set of cells $S_{\text{GC}}=\left\lbrace\left( i,j\right)\right\rbrace$ occupied by the goal region $W_{\text{g}}$. For single goal mission the goal consists in one cell; while for path following, the set consists of all the adjacent cells traversed by the path. 

\begin{figure}[t]
\centering
\includegraphics[width=2.1in]{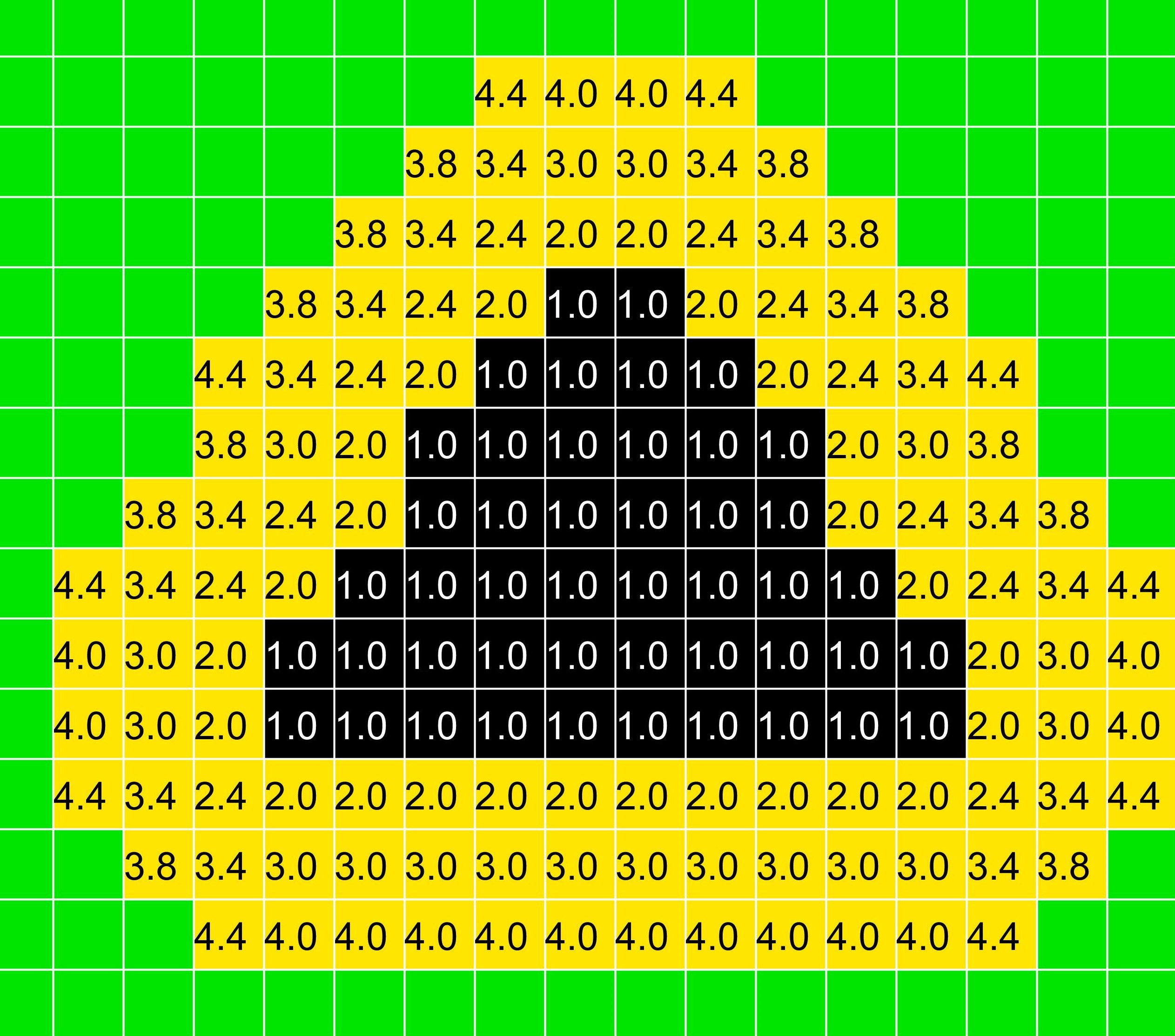}
\caption{Example of Buffer region.}
\label{example_buffer}
\end{figure}

In the first step, a ``complete'' map $M$ is generated by additing the buffer region $W_{\text{buffer}}$ and the goal region $W_{g}$. Like in $BM$, $W_{\text{free}}$ and $W_{\text{obs}}$ are encoded respectively with $0$ and $1$; instead, $W_{\text{buffer}}$ is indicated with the value $-1$, and $W_{\text{g}}$ with $2$. The cells occupied by $W_{\text{buffer}}$ are computed by using the \textit{Brushfire Algorithm} \cite{Choset2005}, in which a grid is used to approximate distances.
\IncMargin{0.3em}
\begin{algorithm}[t]
\SetKwData{map}{M}\SetKwData{bmap}{BM}
\SetKwData{dir}{N}\SetKwData{buffer}{Buffer}
\SetKwData{cost}{Cost}\SetKwData{goals}{$\text{S}_{\text{GC}}$}
\SetKwProg{foreach}{foreach}{ do}{end}
\SetKw{return}{return}
\SetKwInOut{Input}{input}\SetKwInOut{Output}{output}
\Input{bitmap \bmap , goal set \goals}
\Output{complete map \map}
\BlankLine
$\map\leftarrow\bmap$\\
$\buffer\leftarrow\lceil\frac{\alpha r_{\text{min}}}{\text{res}}\rceil$\\
$\cost\leftarrow 1$\\
\While{$\cost<\buffer+1$}{
$\cost\leftarrow \cost+1$\\
\foreach{$\map_{i,j}==0$}{
$\dir_{\text{VH}}\leftarrow \lbrace\map_{i+1,j}, \map_{i,j+1}, \map_{i-1,j},\allowbreak \map_{i,j-1}\rbrace$\\
$\dir_{\text{D}}\leftarrow \lbrace\map_{i+1,j-1}, \map_{i+1,j+1}, \map_{i-1,j+1},\allowbreak \map_{i-1,j-1}\rbrace$\\
\uIf{$\exists \:n\in\dir_{\text{VH}}\vert n\geq 1$}{
$\map_{i,j}\leftarrow \min \lbrace n\in \dir_{\text{VH}}\vert n\geq 1 \rbrace+1$
}
\ElseIf{$\exists\: n\in\dir_{\text{D}}\vert n\geq 1$}{
$\map_{i,j}\leftarrow \min \lbrace n\in \dir_{\text{D}}\vert n\geq 1 \rbrace+1.41$
}
}
}
\foreach{$\map_{i,j}>1$}{
$\map_{i,j}\leftarrow -1$
}
\foreach{$\map_{i,j}\vert \left(i,j\right)\in\goals$}{
$\map_{i,j}\leftarrow 2$
}
\return \map
\caption{Complete Map Genereation}\label{complete_map_gen}
\end{algorithm}\DecMargin{0.3em}
In \cref{complete_map_gen},  $W_{\text{buffer}}$ is computed in the loop from line 4 to line 15 using the 8-point connectivity. The width of the buffer region is computed in Line 2, where \textit{res} is the resolution of the map in terms of cell's width. In \cref{example_buffer}, there is an example of computation of the buffer region. The remaining part of \cref{complete_map_gen} fills the cells occupied by $W_{\text{buffer}}$ and $W_{\text{g}}$.

In the second step the cost map CM is generated applying the wavefront expansion, as shown in  \cref{algo_8_point}, whose differences with respect to \cref{complete_map_gen} is that the expansion starts from cost 2 and all the zero-valued cells are filled. Instead \cref{complete_map_gen} ends once the buffer region has been covered. It is important to notice that in the wavefront expansion performed in \cref{algo_8_point} cells of the buffer region are considered obstacles. 
\IncMargin{0.3em}
\begin{algorithm}[t]
\SetKwData{map}{M}\SetKwData{cmap}{CM}
\SetKwData{dir}{N}
\SetKwProg{foreach}{foreach}{ do}{end}
\SetKw{return}{return}
\SetKwInOut{Input}{input}\SetKwInOut{Output}{output}
\Input{map \map}
\Output{cost map \cmap}
\BlankLine
$\cmap\leftarrow \map$\\
\While{$\exists$ $\cmap_{i,j}==0$}{
\foreach{$\cmap_{i,j}==0$}{
$\dir_{\text{VH}}\leftarrow \lbrace\cmap_{i+1,j}, \cmap_{i,j+1}, \cmap_{i-1,j},\allowbreak \cmap_{i,j-1}\rbrace$\\
$\dir_{\text{D}}\leftarrow \lbrace\cmap_{i+1,j-1}, \cmap_{i+1,j+1}, \cmap_{i-1,j+1},\allowbreak \cmap_{i-1,j-1}\rbrace$\\
\uIf{$\exists\: n \in\dir_{\text{VH}}\vert n \geq 2$}{
$\cmap_{i,j}\leftarrow \min \lbrace n \in \dir_{\text{VH}}\vert n \geq 2 \rbrace+1$
}
\ElseIf{$\exists\: n \in\dir_{\text{D}}\vert n \geq 2$}{
$\cmap_{i,j}\leftarrow \min \lbrace n \in \dir_{\text{D}}\vert n \geq 2 \rbrace+1.41$
}
}
}
\return \cmap
\caption{8-point connectivity WE}\label{algo_8_point}
\end{algorithm}\DecMargin{0.3em}

The cost map CM is then used to generate $\vec{F}_{\text{raw}}\left(x,y\right)$, where in cells of $W_{\text{SS}}$ the vector field is generated using gradient descent, while for cells occupied by $W_{\text{obs}}$ and $W_{\text{buffer}}$, the vectors point towards the closest point of the buffer region's external border.
\begin{figure}[t]
\centering
\begin{subfigure}[t]{0.3\textwidth}
\centering
\includegraphics[width=1\textwidth]{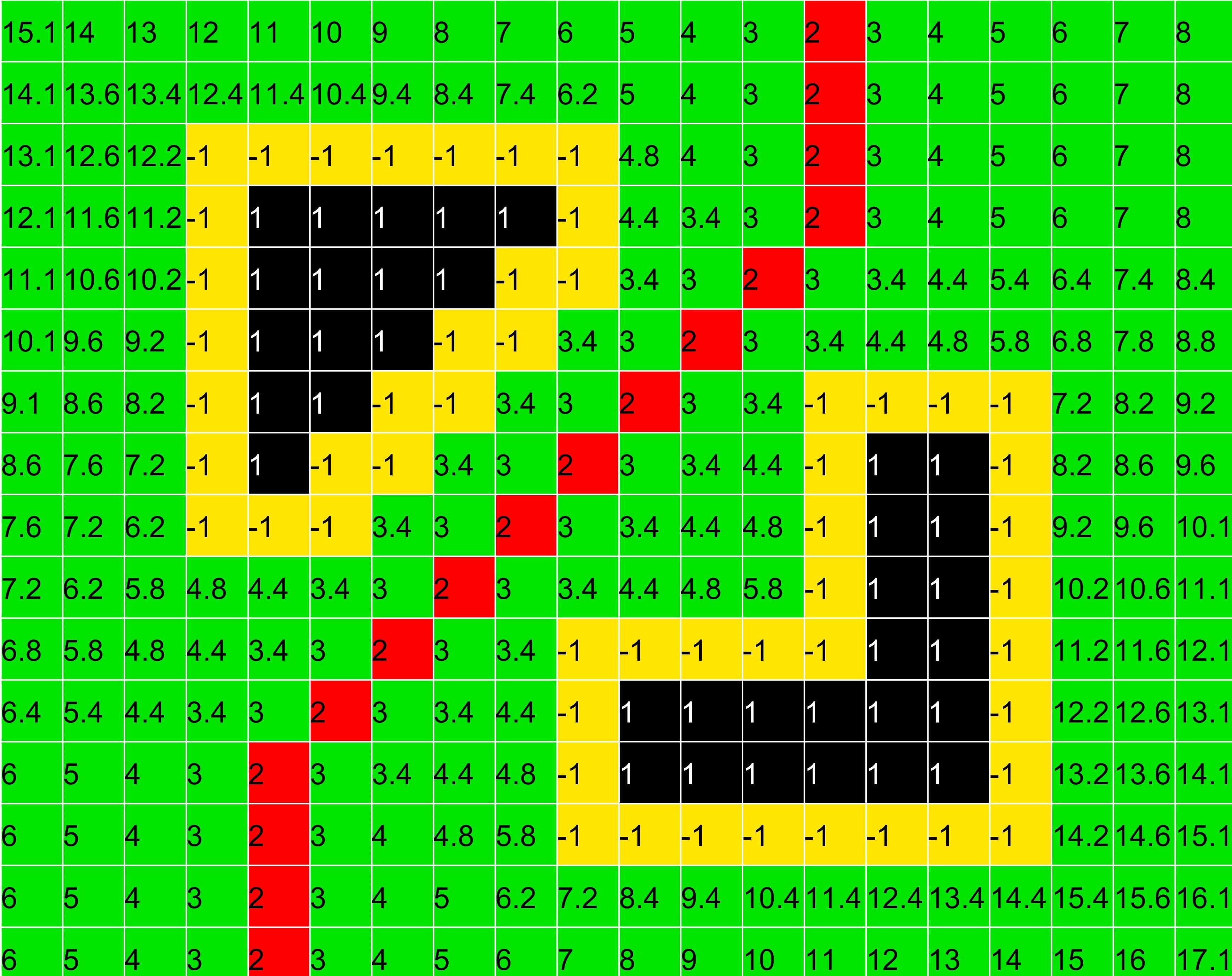}
\caption{ }
\label{comp_map_fig}
\end{subfigure} 
\begin{subfigure}[t]{0.3\textwidth}
\centering
\includegraphics[width=1\textwidth]{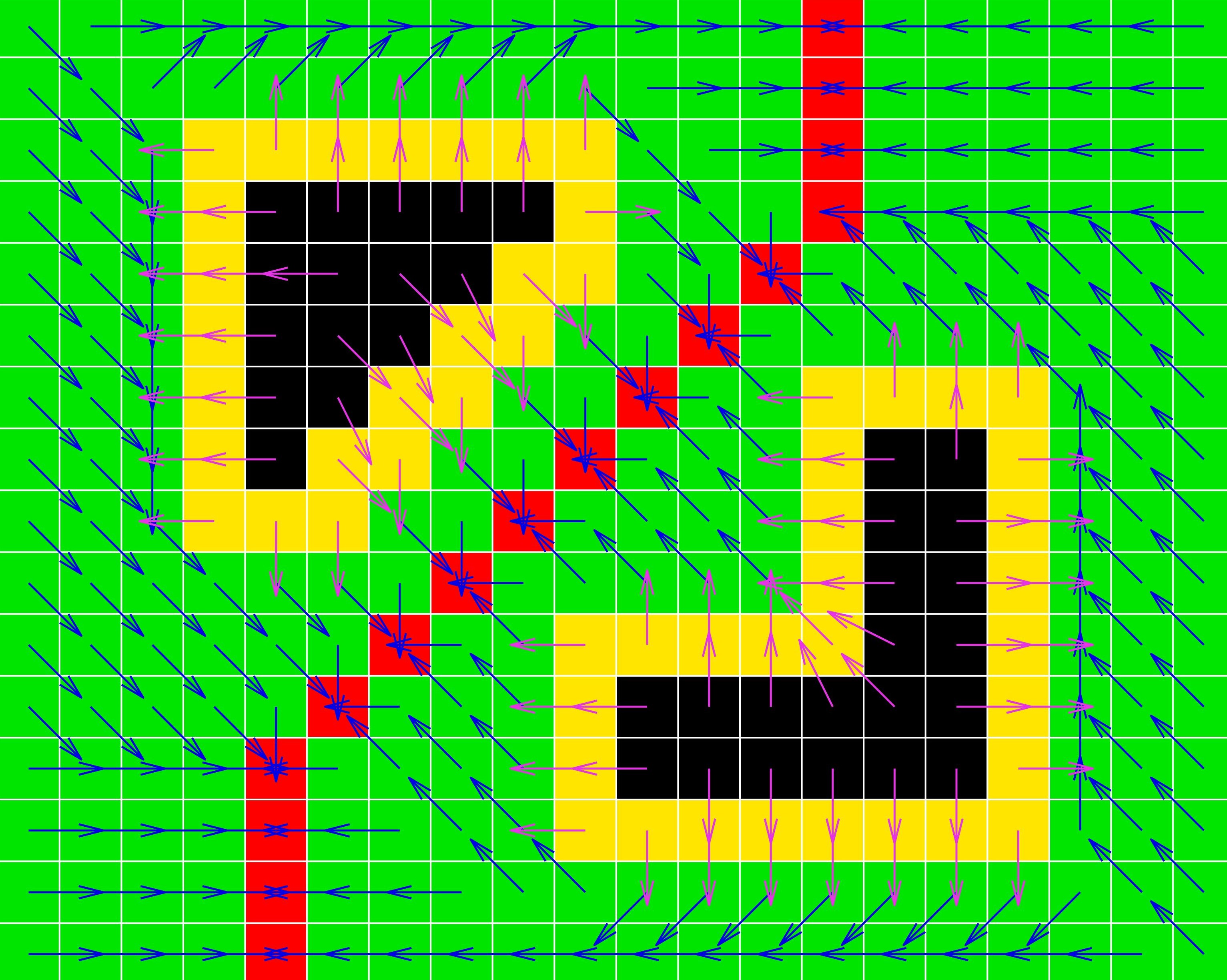}
\caption{ }\label{vec_raw_fig}
\end{subfigure}
\caption{Example of complete map (a) and ``raw'' vector field (b).}
\label{from_complete_to_raw}
\end{figure}

In the generated vector field there are inward flows in proximity of both the path that must be followed and the obstacles' borders. Due to the minimum turning radius constraints, an inward flow leads to oscillations that cause a waste of time and energy. Therefore, it is necessary to mitigate this effect and reduce the presence of oscillations in the path. In our solution, a \text{transition region} is used in proximity of both the target path and the buffer regions' borders. The definition of this region is based on the sliding mode resulting from Filippov's convex method \cite{Filippov2013}, where for a nonlinear system with discontinuous right-hand side
\begin{equation}
\dot{x}\left(t\right)= \begin{cases}
f_{-}\left(t,x\left(t\right)\right), & x\in V_{-}\\
f_{+}\left(t,x\left(t\right)\right), & x\in V_{+}\\
\end{cases}
\label{example_disc_de}
\end{equation}
at switching boundary $\Sigma$, we have:
\begin{equation}
f=af_{+}+(1-a)f_{-},\qquad a\in\left[0,1\right]
\label{example_sliding}
\end{equation}
In our vector field we thicken the boundary obtaining a region having a width proportional to the minimum turning radius. Referring to \cref{example_sliding}, in our case $f_{+}$ is either the vector field tangential to the path or the vector field at the buffer region's outer border. We refer to these two regions with the term ``edge''. Instead, $f_{-}$ is the original vector field in the considered location. We indicate the former with $\vec{F}_{\text{raw}}\left(p_{e}\right)$ and the latter with $\vec{F}_{\text{raw}}\left(p_{c}\right)$, where $p_{e}$ is the position of the closest edge with respect to the considered position $p_{c}$. In our definition, $a$ is a potential function of both the distance from the closest edge and the angle difference between the two vectors. The new vector field $\vec{F}_{t}$ is obtained as follows:
\begin{equation}
\vec{F}_{t}\left(p_{c}\right)= \begin{cases}
\vec{F}_{\text{raw}}\left(p_{c}\right), &\Vert p_{c}-p_{e}\Vert>\sigma_{*}r_{\text{min}}\\
R\left(\theta\left(p_{c},p_{e}\right)\right)\vec{F}_{\text{raw}}\left(p_{c}\right),&\Vert p_{c}-p_{e}\Vert\leq\sigma_{*}r_{\text{min}}\\
\end{cases}
\label{eq_transition}
\end{equation}
\begin{equation}
R\left(\theta\right)=
\begin{bmatrix}
    \cos\theta &-\sin\theta\\
    \sin\theta &\cos\theta\\
\end{bmatrix}
\label{rot_matrix}
\end{equation}
\begin{equation}
\theta\left(p_{c}, p_{e}\right)= \begin{cases}
(1-a)\Delta\theta_{ce}, &\angle\theta_{p_{e}}<\angle\theta_{p_{c}}\\
(1-a)\Delta\theta_{ce}, &\angle\theta_{p_{e}}\geq\angle\theta_{p_{c}}
\end{cases}
\label{eq_theta_rot}
\end{equation}
\begin{equation}
a=\mu_{*}\frac{\Vert p_{c}-p_{e}\Vert}{\sigma_{*}r_{\text{min}}}\frac{\Delta\theta_{ce}}{\pi}
\label{a_formula}
\end{equation}
\begin{equation}
\Delta\theta_{ce}=\cos^{-1}\left(\vec{F}_{\text{raw}}\left(p_{c}\right)\cdot\vec{F}_{\text{raw}}\left(p_{e}\right)\right)
\label{delta_ce_formula}
\end{equation}
\begin{equation}
p_{*}=\left(x_{*}, y_{*}\right)
\label{p_definition}
\end{equation}

The tuning parameters for the transition regions are: $\mu_{p}$, $\sigma_{p}$, $\mu_{b}$, $\sigma_{b}$ where $p$ stands for path and $b$ for border. The value of $\mu\in\left(0,1\right]$ is used to tune the ``speed'' of the rotation (i.e. the smaller the value, the sooner the flow becomes parallel to the edge). Instead $\sigma\in\left[1,2\right]$ regulates the width of the transition region.

To further reduce the presence of oscillations, in the final step of our method, we exploit a technique that is widely used in image processing. In particular, we use a Gaussian filter \cite{Szeliski2010}, which is appealing for our application because it removes high frequency components, therefore in our case it reduces abrupt changes in the flow's direction. A 2-D Gaussian is defined as follows:
\begin{equation}
G\left(x, y, \sigma\right)=\frac{1}{2\pi\sigma^{2}}e^{-\frac{x^{2}+y^{2}}{2\sigma^{2}}}
\label{eq_gauss}
\end{equation}
A Gaussian kernel is a discrete approximation (i.e. a matrix) of \cref{eq_gauss}. In our filter, we employ a kernel with a variable size, which depends upon the standard deviation $\sigma$:
\begin{equation}
n=2\lceil 2\sigma\rceil+1
\label{gauss_dim}
\end{equation}
The final vector field $\vec{F}\left(x, y\right)$ is obtained by performing the convolution between each component of $\vec{F}_{t}\left(x, y\right)$ and the Gaussian kernel:
\begin{equation}
\vec{F}\left(x,y\right)=\left[ \vec{F}_{tx}\left(x,y\right)*G_{\sigma}, \vec{F}_{ty}\left(x,y\right)*G_{\sigma} \right]^{T}
\label{filtering}
\end{equation}

\begin{figure}[t]
\centering
\begin{subfigure}[t]{0.24\textwidth}
\centering
\includegraphics[width=1\textwidth]{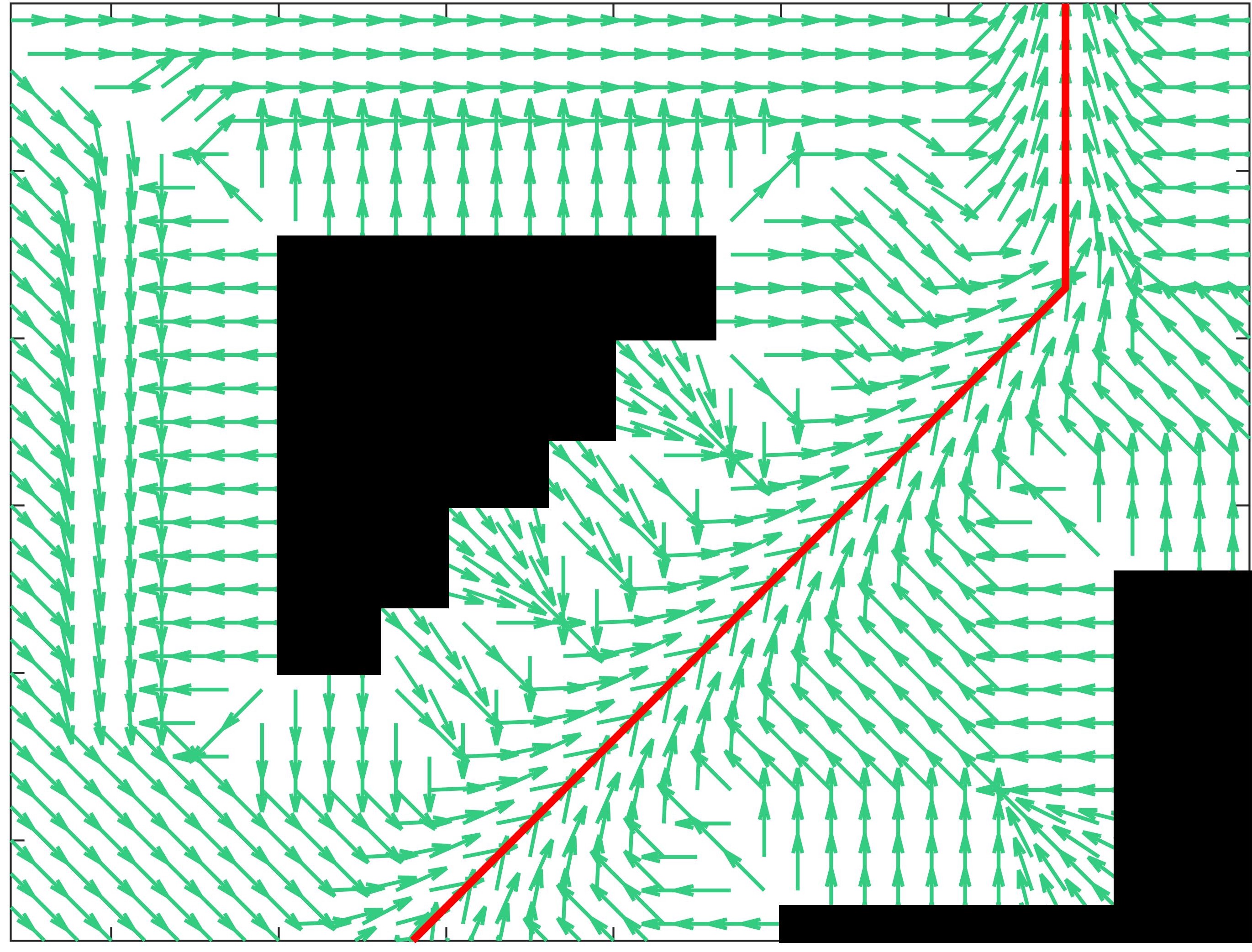}
\caption{ }
\label{fig_not_smooth}
\end{subfigure} 
\begin{subfigure}[t]{0.24\textwidth}
\centering
\includegraphics[width=1\textwidth]{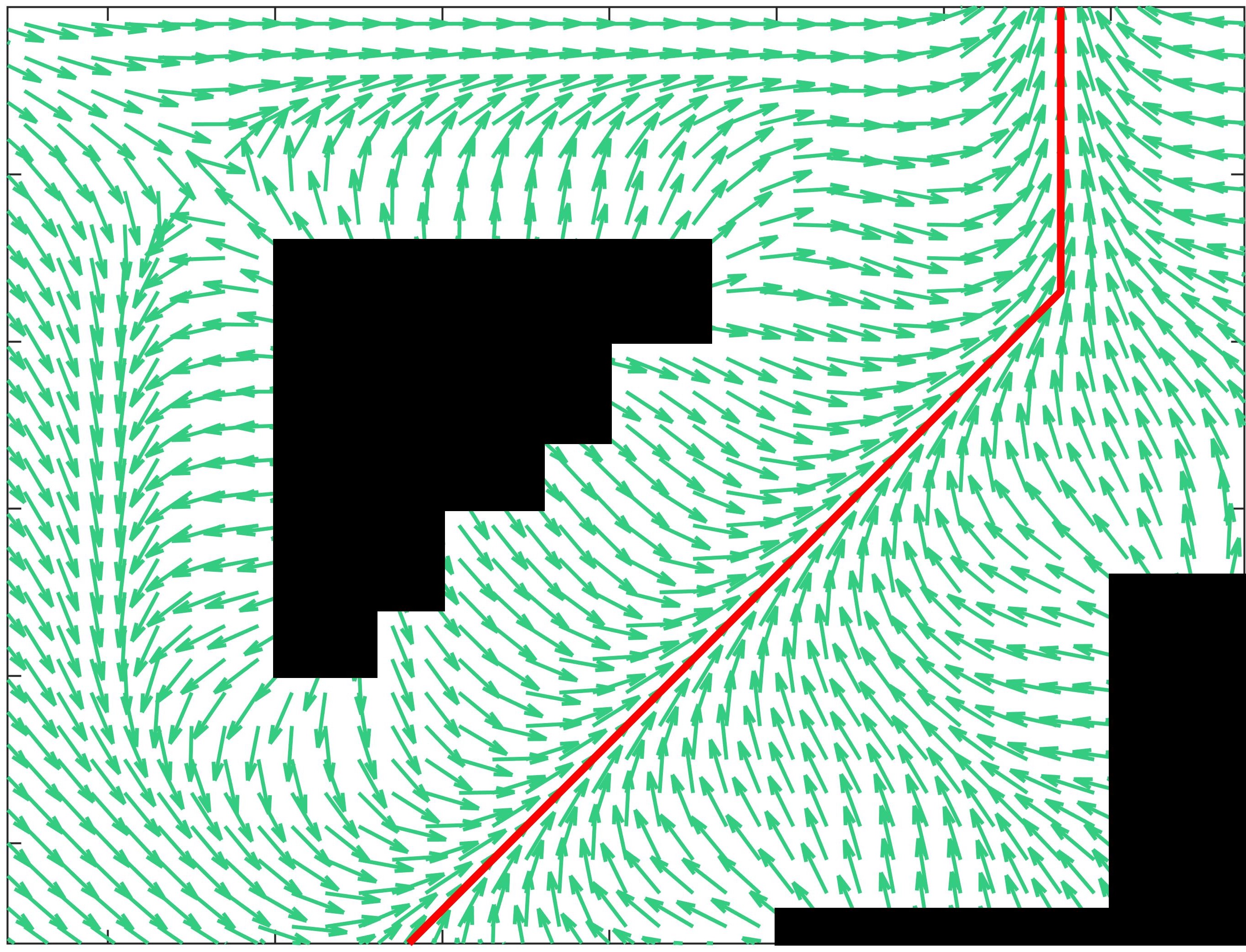}
\caption{ }\label{fig_smooth}
\end{subfigure}
\caption{Example of non-smoothed vector field (a) and smoothed vector field (b).}
\label{smoothing}
\end{figure}

\section{Simulation Results}

\begin{figure}[t]
\centering
\begin{subfigure}[t]{0.3\textwidth}
\centering
\includegraphics[width=1\textwidth]{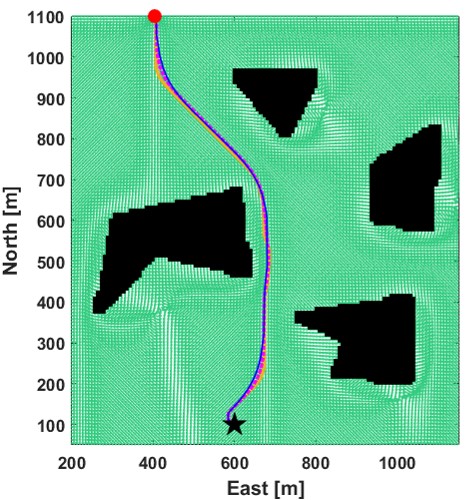}
\caption{ }
\label{single_goal}
\end{subfigure} 
\begin{subfigure}[t]{0.35\textwidth}
\centering
\includegraphics[width=1\textwidth]{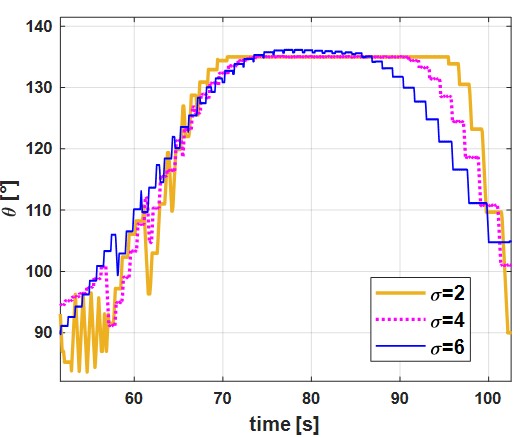}
\caption{ }\label{single_goal_theta}
\end{subfigure}
\caption{Example of single goal mission (a) and corresponding $\theta$ (b) for $\sigma=2$ (solid yellow line), $\sigma=4$ (dotted purple line), $\sigma=6$ (solid blue line).}
\label{mission_1}
\end{figure}
\begin{figure}[t]
\centering
\begin{subfigure}[t]{0.3\textwidth}
\centering
\includegraphics[width=1\textwidth]{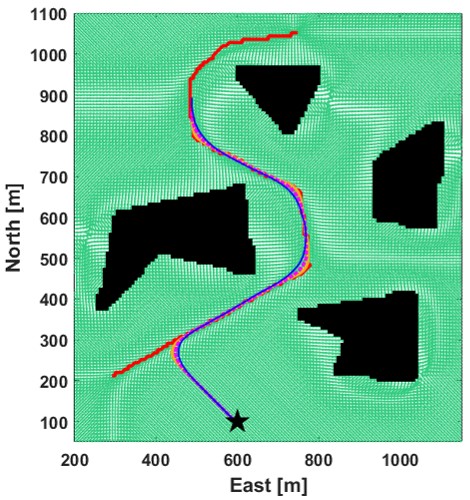}
\caption{ }
\label{path_following}
\end{subfigure} 
\begin{subfigure}[t]{0.35\textwidth}
\centering
\includegraphics[width=1\textwidth]{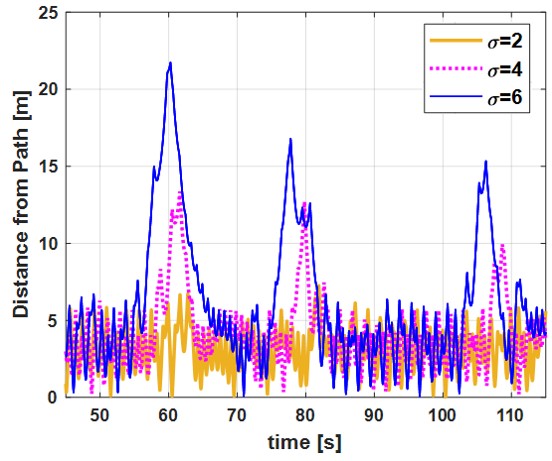}
\caption{ }\label{path_following_distance}
\end{subfigure}
\caption{Example of path following mission (a) and corresponding distance from path (b) for $\sigma=2$ (solid yellow line), $\sigma=4$ (dotted purple line), $\sigma=6$  (solid blue line).}
\label{mission_2}
\end{figure}
In this section, we illustrate simulation results relative to two scenarios. In the first one, the vehicle must reach a single goal location; instead, in the second one it must follow a predefined path. These results were obtained simulating the kinematic model of a vehicle moving in a cluttered environment with obstacles having irregular shapes. The vehicle was assumed to travel with a linear speed $v=10 \frac{m}{s}$ and a minimum turning radius $r_{\text{min}}= 20m$. 

In the examples illustrated in \cref{mission_1} and \cref{mission_2}, the feedback motion plan $\pi$ was based on a discrete vector field having a resolution of $8m$. In both examples, we compare the resulting trajectories for different values of $\sigma$ used in the Gaussian filter.
In \cref{single_goal}, there are paths obtained with three different values of $\sigma$, while \cref{single_goal_theta} shows the vehicle's orientation for the three cases. As expected, the increment of the parameter $\sigma$ leads to smoother transitions in the vehicle's heading angle. In fact, with higher $\sigma$ we obtain a smoother vector field. 

For the case of path following, \cref{path_following} illustrates the resulting paths, while the chart in \cref{path_following_distance} shows the distance of the vehicle from the predefined path that must be followed. We can notice that with greater values of $\sigma$ there are higher deviations from the predefined path. This is also expected because the Gaussian filter removes high-frequency components limiting changes in the vector field. Therefore, sharp corners in the path are smoothed.

\begin{figure}[t]
\centering
\begin{subfigure}[t]{0.4\textwidth}
\centering
\includegraphics[width=1\textwidth]{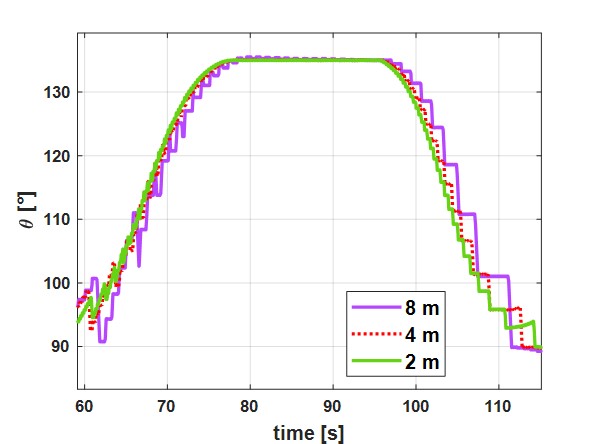}
\caption{ }\label{single_goal_theta_res}
\end{subfigure}
\begin{subfigure}[t]{0.4\textwidth}
\centering
\includegraphics[width=1\textwidth]{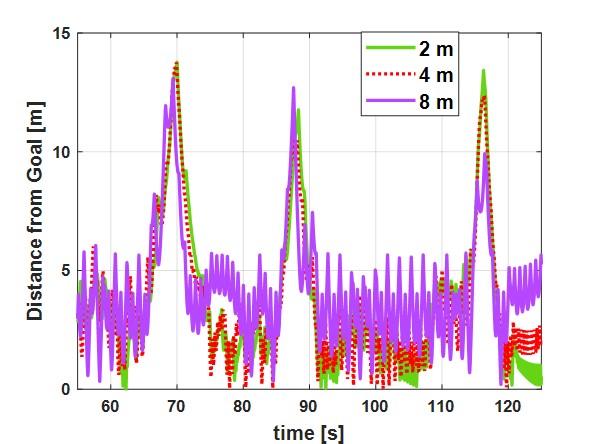}
\caption{ }\label{path_following_distance_res}
\end{subfigure}
\caption{Heading angle in single goal mission (a) and distance error in a path following mission (b) for $res=2m$ (solid green line), $res=4m$ (dotted red line), $res=8m$  (solid purple line).}
\label{mission_res}
\end{figure}

While in \cref{mission_1} and \cref{mission_2} we compared the cases of different values of $\sigma$, in \cref{mission_res} we compare the cases obtained with different resolutions of the grid-map. In particular, in this case, the resolutions of $2m$, $4m$, and $8m$ were tested for both the single goal mission and the path following mission illustrated in \cref{single_goal} and \cref{path_following}. In this case, in order to have the same level of "smoothness", we had to use a value of sigma proportional to the grid resolution. Therefore, since for $8m$ we used $\sigma=4$, for the cases of $4m$ and $2m$, we used $\sigma=8$ and $\sigma=16$, respectively . As expected, with the resolution of $8m$ we obtain the least smooth path in the single goal mission and the highest average error in the path following. However, it is interesting to notice that while the average error is higher with the $8m$ resolution, the spikes obtained at the corners are the same as those obtained with $2m$ and $4m$. 

The examples discussed above illustrate the effectiveness of the method. Nevertheless, they also indicate that further research needs to be conducted on this method to define the criteria for optimal tuning of the parameters. In particular, we have seen that the same parameter needs to be tuned differently based on the type of mission. In fact, in the first case, a higher $\sigma$ is desirable to limit chattering. On the other hand, for a path following mission, a lower  $\sigma$ is preferable for more precise tracking of the predefined path. Finally, the tuning of the grid-map resolution versus the quality of the solution is another relationship that must be investigated. In fact, it is always desirable to use the minimum number of cells in order to reduce the computational time. Nevertheless, the number of cells must be high enough to have an acceptable solution.

\section{Conclusion}
In this paper, we formulated a feedback motion plan based on the extension of the \textit{wavefront expansion} to the case of bounded curvature constraints. The approach can be used for both single goal missions and path following missions in the presence of obstacles having arbitrary shape.

The effectiveness of the method was demonstrated using simulated examples. The obtained results encourage us to focus our future work on the derivation of optimal tuning parameters accounting for both vehicle kinematics and environmental constraints. For instance, in the examples provided in this paper, we could see that the tuning of $\sigma$ in the Gaussian filter must be different for the case of single goal mission and path following. 
  
\bibliography{mybibfile}
\bibliographystyle{IEEEtran}

\end{document}